%% file: main.tex
\documentclass{article}

\usepackage{wrapfig, blindtext} 

\usepackage[preprint,nonatbib]{neurips_2024}

\usepackage[utf8]{inputenc} 
\usepackage[T1]{fontenc}    
\usepackage{hyperref}       
\usepackage{url}            
\usepackage{booktabs}       
\usepackage{amsfonts}       
\usepackage{nicefrac}       
\usepackage{microtype}      
\usepackage{xcolor}         
\usepackage{enumitem}
\usepackage{amssymb}
\usepackage{pifont}
\usepackage{lipsum}
\usepackage{graphicx} 
\usepackage{framed}
\usepackage{amsmath}
\usepackage{cleveref}

\hypersetup{
    colorlinks=true,
    linkcolor=magenta,
}

\newcommand\blfootnote[1]{%
  \begingroup
  \renewcommand\thefootnote{}\footnote{#1}%
  \addtocounter{footnote}{-1}%
  \endgroup
}
\begin{document}
\input{sections/predefine}
\title{\papertitle}
\author{%
  Yiming Zhang\textsuperscript{\rm 1}\quad
  Yicheng Gu\textsuperscript{\rm 2}\quad
Yanhong Zeng\textsuperscript{\rm 1,$\dagger$} \quad
Zhening Xing\textsuperscript{\rm 1}\\
\textbf{Yuancheng Wang}\textsuperscript{\rm 2}\quad
\textbf{Zhizheng Wu}\textsuperscript{\rm 2,1}\quad
\textbf{Kai Chen}\textsuperscript{\rm 1,$\dagger$}
\\
\vspace{-0.6em}\\
$^1$Shanghai Artificial Intelligence Laboratory \quad
$^2$Chinese University of Hong Kong, Shenzhen\\ \\
\textbf{Homepage: }\href{https://foleycrafter.github.io/}{https://foleycrafter.github.io/}\\
\vspace{-1.5em}\\
}
\blfootnote{\textsuperscript{$\dagger$} denotes corresponding author.} 
\vspace{-20pt}

\maketitle
\input{sections/00_abstract}
\input{sections/01_intro}

\input{sections/02_related}

\input{sections/03_approach}
\input{sections/04_experiment}

\input{sections/05_conclusion}

\bibliographystyle{plain} 
\bibliography{reference}

\appendix

\input{sections/06_appendix}

\end{document}

%% file: sections/predefine.tex
\newcommand{\ie}{\textit{i.e.,}} 
\newcommand{\eg}{\textit{e.g.,}} 
\newcommand{\etc}{\textit{etc.}} 
\newcommand{\etal}{\textit{etal.}}

\newcommand{\cmark}{\ding{51}}%
\newcommand{\xmark}{\ding{55}}%
\newcommand{\todomark}{$\square$}
\newcommand{\done}{\rlap{$\square$}{\raisebox{2pt}{\large\hspace{1pt}\cmark}}%
\hspace{-2.5pt}}
\newcommand{\wontfix}{\rlap{$\square$}{\large\hspace{1pt}\xmark}}

\newcommand{\modelname}{FoleyCrafter}

\newcommand{\papertitle}{
{
    \vspace{-10pt}
    \raisebox{-0.3\height}{\includegraphics[width=0.1\linewidth]{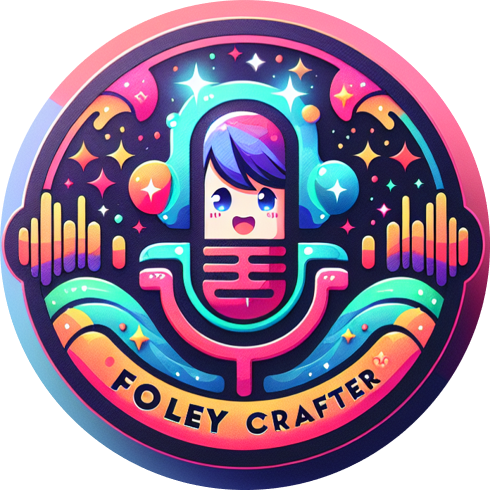}}
        {\modelname: Bring Silent Videos to Life with Lifelike and Synchronized Sounds}
    }
}

\newcommand{\semanticmodule}{semantic adapter}

\newcommand{\temporalmodule}{temporal controller}

\newcommand{\yh}[1]{{\color{red}#1}}
\newcommand{\highlight}[1]{{\color{red}#1}}
\newcommand{\todo}[1]{{\color{lightgray}#1}}


%% file: sections/00_abstract.tex

\begin{abstract}
We study Neural Foley, the automatic generation of high-quality sound effects synchronizing with videos, enabling an immersive audio-visual experience.
Despite its wide range of applications, existing approaches encounter limitations when it comes to simultaneously synthesizing high-quality and video-aligned (\ie, semantic relevant and temporal synchronized) sounds.
To overcome these limitations, we propose \modelname, a novel framework that leverages a pre-trained text-to-audio model to ensure high-quality audio generation. \modelname\ comprises two key components: the \semanticmodule\ for semantic alignment and the \temporalmodule\ for precise audio-video synchronization. The \semanticmodule\ utilizes parallel cross-attention layers to condition audio generation on video features, producing realistic sound effects that are semantically relevant to the visual content. Meanwhile, the \temporalmodule\ incorporates an onset detector and a timestamp-based adapter to achieve precise audio-video alignment.
One notable advantage of \modelname\ is its compatibility with text prompts, enabling the use of text descriptions to achieve controllable and diverse video-to-audio generation according to user intents.
We conduct extensive quantitative and qualitative experiments on standard benchmarks to verify the effectiveness of \modelname. Models and codes are available at \href{https://github.com/open-mmlab/FoleyCrafter}{https://github.com/open-mmlab/FoleyCrafter}.
\end{abstract}

%% file: sections/01_intro.tex
\section{Introduction}

Foley, a key element in film and video post-production, adds realistic and synchronized sound effects to silent media \cite{foleywiki}. These sound effects are the unsung heroes of cinema and gaming, enhancing realism, impact, and emotional depth for an immersive audiovisual experience.
Traditionally, skilled Foley artists painstakingly create, record, and process sound effects in specialized studios, making it a labor-intensive and time-consuming process \cite{ament2014foley}.
Despite advancements in recent video-to-audio generation, achieving Neural Foley, which requires synthesizing high-quality, video-aligned sounds that are semantically related and temporally synchronized with the videos, remains challenging \cite{luo2024diff}.

\begin{figure}[h]
    \vspace{-2mm}
    \centering
    \includegraphics[width=\linewidth]{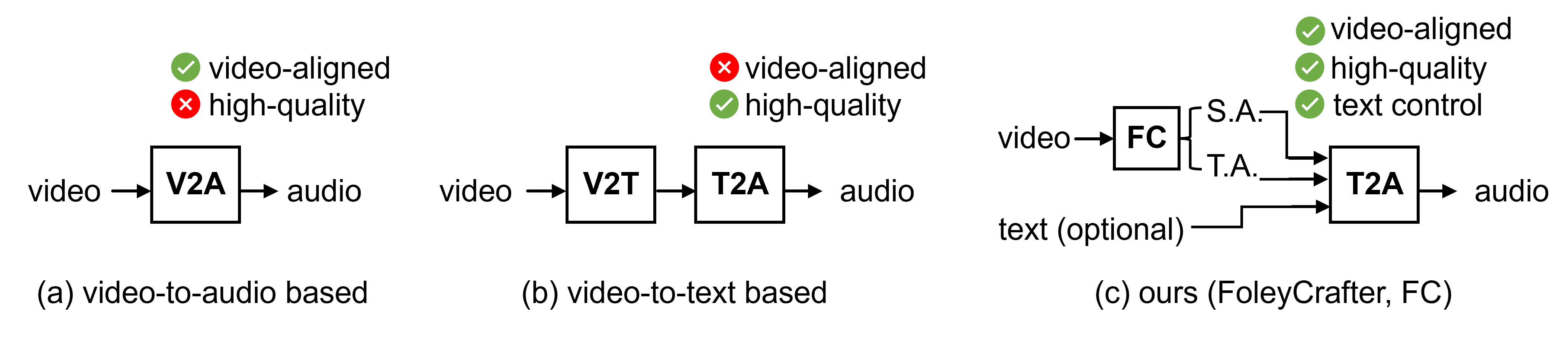}
    \vspace{-5mm}
    \caption{(a) (Video-to-audio) V2A methods struggle with audio quality due to noisy training data, while (b) video-to-text (V2T) methods encounter difficulties in producing synchronized sounds. Our model, \modelname\ (FC), integrates a learnable module into a pre-trained Text-to-Audio (T2A) model to ensure audio quality while enhancing video-audio alignment with the supervision of audios.}
    \label{fig:comp}
\end{figure}



State-of-the-art approaches for Neural Foley in video-to-audio generation can be categorized into two main groups, as illustrated in \cref{fig:comp}. 
The first group involves training a video-to-audio generative model on a large-scale paired audio-video dataset \cite{chen2020vggsound, iashin2021taming, luo2024diff, sheffer2023hear}. However, the audio quality of such datasets crawled from the Internet can be subpar, with issues like noise and complex environmental sounds recorded in the wild, which hinder the production of high-quality sounds \cite{wang2024v2a,xie2024sonicvisionlm}.
To address this, the second group of approaches (\cref{fig:comp}-(b)) adopts a two-stage process. They first translate video into text using video captioning or embedding mapping techniques and then employ a pre-trained text-to-audio model \cite{wang2024v2a, xie2024sonicvisionlm, xing2024seeing}. Leveraging the well-trained text-to-audio generator, these methods achieve impressive sound quality. Nonetheless, effectively bridging the gap between video and text while preserving nuanced details is challenging. As a result, these methods may produce unsynchronized sounds due to the suboptimal translated text conditions.

To achieve both high-quality and video-aligned sound generation, we present \modelname, which breathes life into silent videos with realistic and synchronized sound effects.
As depicted in \cref{fig:comp}-(c), the core of \modelname\ is an innovative pluggable module that can be integrated with a pre-trained text-to-audio (T2A) model, optimized with the supervision of audios.
Specifically, \modelname\ comprises two main components: a \semanticmodule\ for semantic alignment and a \temporalmodule\ for temporal synchronization.
The \semanticmodule\ introduces parallel cross-attention layers into the backbone of the T2A model. It takes as input the extracted video features, allowing \modelname\ to generate audio conditioned on the video without relying on explicit text.
The \temporalmodule, on the other hand, is engineered to refine temporal synchronization. 
The \temporalmodule\ features an onset detector and a timestamp-based adapter. The onset detector is trained to predict sound occurrences and silences from videos. The timestamp-based adapter then refines this by aligning audio generation with the predicted timestamps, ensuring sound and silence are synchronized with the video's temporal flow.
During training, we train the \semanticmodule\ and \temporalmodule\ with video-audio correspondent data, while fixing the text-to-audio base model to preserve its established audio generation quality. After training, \modelname\ can generate high-quality sounds for videos with semantic and temporal alignment in a flexible and controllable way. 

We conduct extensive experiments to evaluate the performance of \modelname\ in terms of audio quality and video alignment, both semantically and temporally. Our experiments include quantitative analysis, qualitative comparison, and user studies, all of which demonstrate that \modelname\ has achieved state-of-the-art results. Additionally, we have showcased the controllability of \modelname\ through text prompts, allowing for a more fine-grained and versatile application of the model.
Our main contributions can be summarized as follows:
\begin{itemize}
    \item We present a novel Neural Foley framework that generates high-quality, video-aligned sound effects for silent videos, while also offering fine-grained control through text prompts.
    \item To ensure both semantic and temporal alignment, we design a \semanticmodule\ and a \temporalmodule, significantly improving video alignment.
    \item We validate the effectiveness of \modelname\ through extensive experiments, including quantitative and qualitative analyses. Our results show that \modelname\ achieves state-of-the-art performance on commonly used benchmarks.
\end{itemize}

%% file: sections/02_related.tex
\section{Related Work}

\paragraph{Diffusion-based Audio Generation.}Latent diffusion models have significantly advanced audio generation \cite{liu2023audioldm,liu2023audioldm2,rombach2022high}. AudioLDM pioneers open-domain text-to-audio generation using a latent diffusion model \cite{liu2023audioldm,liu2023audioldm2}. Tango improves text-to-audio generation with an instruction-tuned LLM FLAN-T5 \cite{chung2024scaling} as the text encoder \cite{ghosal2023text}. Make-an-Audio tackles complex audio modeling using spectrogram autoencoders instead of waveforms \cite{huang2023make}. Li et al. conduct comprehensive ablation studies to explore effective designs and set a new state-of-the-art with the proposed Auffusion \cite{xue2024auffusion}.
In this paper, we introduce \modelname, a module that extends state-of-the-art text-to-audio generators to support video-to-audio generation while preserving the original text-to-audio controllability.


\paragraph{Video-to-Audio Generation.} 
Foley artistry is a crucial audio technique that enhances the viewer's auditory experience by creating and recording realistic sound effects that synchronize with visual content \cite{foleywiki}. 
Early Neural Foley models mainly focus on generating sounds tailored to a specific genre or a narrow spectrum of visual cues, underscoring the potential of deep learning to innovate sound effect creation for videos \cite{chen2018visually,chen2020generating,owens2016visually,zhou2018visual}. 
Despite recent advancements in large-scale generative models \cite{huang2023make,liu2023audioldm}, generating high-quality and visually synchronized sounds for open-domain videos remains a challenge \cite{dong2023clipsonic,du2023conditional,luo2024diff,mo2024text,tang2024any,wang2024v2a}.

State-of-the-art video-to-audio approaches can be categorized into two groups. The first group focuses on training video-to-audio generators from scratch \cite{iashin2021taming,luo2024diff,sheffer2023hear}. Specifically, 
SpecVQGAN \cite{iashin2021taming} employs a cross-modal Transformer \cite{vaswani2017attention} to auto-regressively generate sounds from video tokens. Im2Wav \cite{sheffer2023hear} conditions an autoregressive audio token generation model using CLIP features, while Diff-Foley \cite{luo2024diff} improves semantic and temporal alignment through contrastive pre-training on aligned video-audio data. However, these methods are limited by the availability of high-quality paired video-audio datasets. 
An alternative approach is to utilize text-to-audio generators for video Foley. Xing et al. \cite{xing2024seeing} introduce an optimization-based method with ImageBind \cite{girdhar2023imagebind} for video-audio alignment, while SonicVisionLM \cite{xie2024sonicvisionlm} generates video captions for text-to-audio synthesis. Wang et al. note the limitations of caption-based methods and propose V2A-Mapper to translate visual embeddings to text embedding space \cite{wang2024v2a}. Nevertheless, effectively bridging the gap between video and text while preserving fine-grained temporal cues remains a significant challenge.
In contrast, we introduce \modelname, integrating a learnable module into text-to-audio models with end-to-end training, enabling a high-quality, video synchronized and high-controllable Foley.

%% file: sections/03_approach.tex
\newcommand{\figframework}{
\begin{figure}[t]
    \centering
    \includegraphics[width=\textwidth]{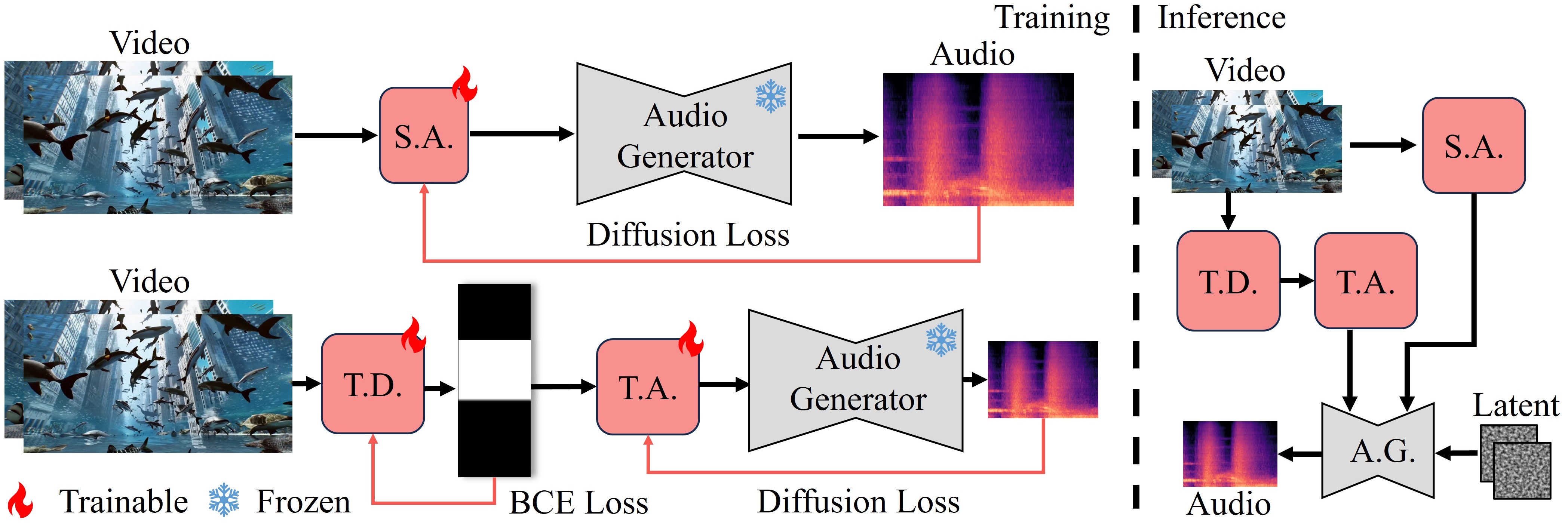}
    \caption{
    \textbf{The overview of \modelname.} \modelname\ is built upon a pre-trained text-to-audio (T2A) generator, ensuring high-quality audio synthesis. 
    It comprises two main components: the \semanticmodule\ (S.A.) and the \temporalmodule, which includes a timestamp detector (T.D.) and a temporal adapter (T.A.).
    Both the \semanticmodule\ and the \temporalmodule\ are trainable modules that take videos as input to synthesize audio, with audio supervision for optimization. The T2A model remains fixed to maintain its established capability for high-quality audio synthesis.
    }
    \label{fig:framework}
\end{figure}
}

\newcommand{\figsemantic}{
\begin{figure}[t]
    \centering
    \includegraphics[width=\textwidth]{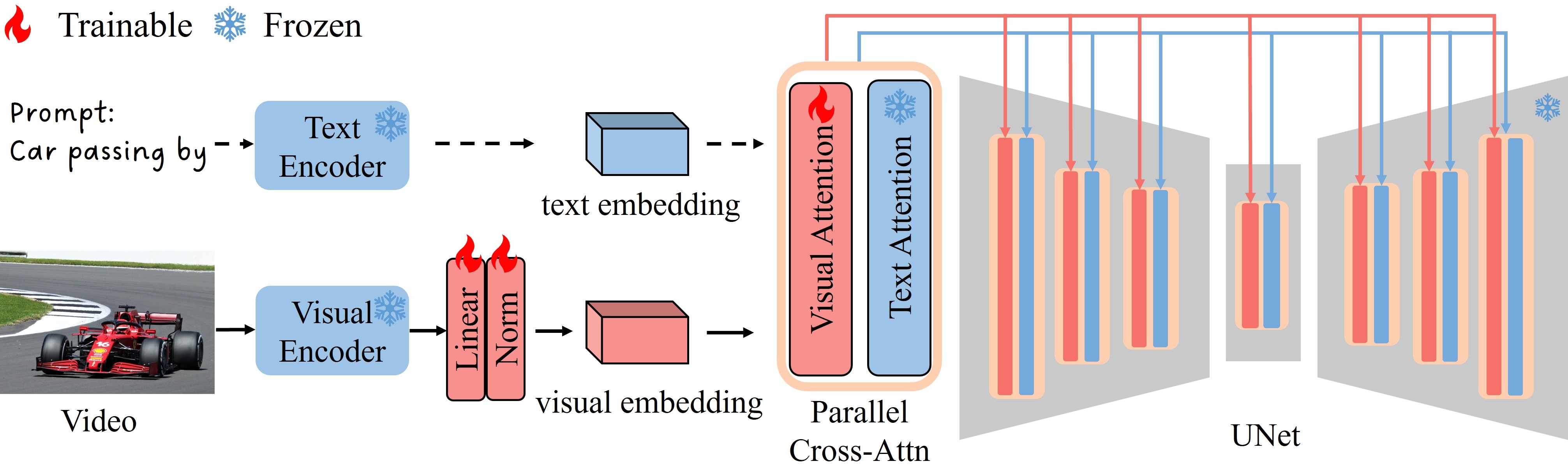}
    \caption{
    \textbf{The overview of \semanticmodule}. 
    Semantic adapter employs a pre-trained visual encoder with several learnable layers to extract video embeddings that align better with the text-to-audio generator. Then, it integrates trainable visual-cross attention mechanisms alongside text-based ones, ensuring semantic alignment with the video without compromising text-to-audio generation.
    }
    \label{fig:semantic}
\end{figure}
}

\newcommand{\figtemporal}{
\begin{figure}[t]
    \centering
    \includegraphics[width=\textwidth]{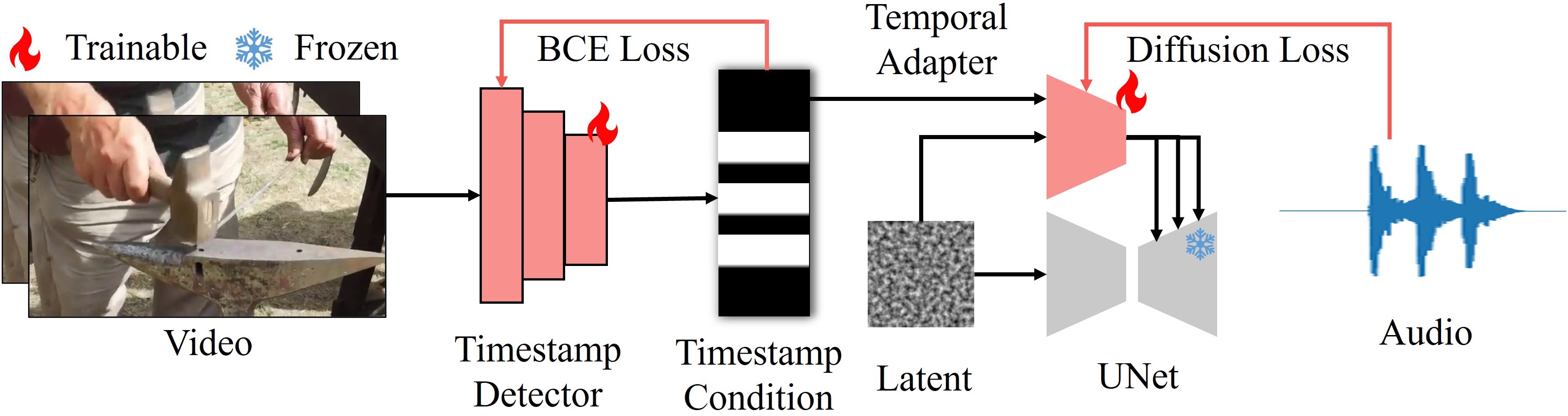}
    \caption{\textbf{The overview of the \temporalmodule.} 
    The temporal module consists of a timestamp detector and a temporal adapter for improved video-audio synchronization. The timestamp detector predicts sound and silence labels based on the video, optimized using ground truth audio event timestamps. The temporal adapter, initialized from UNet encoder blocks, encodes the timestamp condition and injects synchronization information into the UNet decoder.
    }
    \label{fig:temporal}
\end{figure}
}

\section{Approach}
\label{sec:app}
In this section, we introduce the framework of \modelname. We introduce related preliminaries about Audio Latent Diffusion Models (ALDMs) \cite{liu2023audioldm,liu2023audioldm2} and conditioning mechanisms in \cref{subsec:prelimi}. We then delve into the key components of \modelname\ in \cref{subsec:model}. The \semanticmodule\ generates audio based on visual cues and text prompts, while the \temporalmodule\ improves temporal synchronization with the input video. We also outline the training process for each component and explain how \modelname\ can be used to generate foley for silent videos in \cref{subsec:train}.

\subsection{Preliminaries} 
\label{subsec:prelimi}
\paragraph{Audio Latent Diffusion Model.}
The latent diffusion model (LDM) has achieved remarkable advancements in text-to-audio generation, as demonstrated by recent studies \cite{ghosal2023text,liu2023audioldm,liu2023audioldm2,xue2024auffusion}. In this model, the audio waveform is initially transformed into a mel-spectrogram representation. Subsequently, a variational autoencoder (VAE) encodes the mel-spectrogram into a latent representation denoted as $z$. The LDM's UNet is trained on this latent space to generate $z$ by denoising normally distributed noise $\epsilon$. The predicted latent $z$ is then reconstructed by the VAE into a mel-spectrogram, which is finally transformed into a waveform using a vocoder.

A latent diffusion model consists of two main processes: the diffusion process and the denoising process. In the diffusion process, a clean latent representation $z$ undergoes step-by-step noise addition until it reaches an independently and identically distributed noise. Such a process is denoted as,
\begin{equation}
    z_t = \sqrt{\bar{\alpha_t}} z_0 + \sqrt{1 - \bar{\alpha_t}} \epsilon, \epsilon \sim \mathcal{N}(0, I)
    \label{equ:diffusion}
\end{equation}
where $\bar{\alpha_t}$ is the noise strength at $t$ timestep. 
The UNet is trained to estimate the added noise at a given timestep $t$ using the following optimization objective:
\begin{equation}
    \mathcal{L} = \mathbb{E}_{x,\epsilon\sim N(0, 1), t, c} \left[\Vert \epsilon - \epsilon_{\theta}(z_t, t, c)  \Vert \right]
    \label{equ:loss_sd2}
\end{equation}
where $x$ represents the mel-spectrogram in the ALDM, $z_t$ corresponds to the latent representation of the mel-spectrogram at timestep $t$, and $c$ denotes the condition information. 

\paragraph{Conditioning Mechanisms.}
There are two kinds of condition mechanisms mainly used in ALDM, \ie\ MLP-based mechanism \cite{ghosal2023text,liu2023audioldm} and cross-attention-based mechanism \cite{xue2024auffusion,liu2023audioldm2}. In the MLP-based mechanism, the time step is mapped to a one-dimensional embedding and concatenated with the text embedding as the conditioning information. This one-dimensional condition vector is then merged with the UNet's feature map through MLP layers.
In contrast, the cross-attention-based mechanism utilizes a cross-attention operation in each block of the UNet backbone. This mechanism demonstrates improved alignment with conditions and allows for more flexible and fine-grained controllable generation. It has been widely adopted in recent works \cite{liu2023audioldm2,xue2024auffusion}. The cross-attention mechanism can be represented as follows:
\begin{align}
    Attention(Q, K, V) &= softmax\left(\frac{QK^T}{\sqrt{d}}\right) \cdot V,\\
    \text{where } &Q = W_{Q} \cdot \varphi(z_t), \quad
                  K = W_{K} \cdot \tau(c), \quad
                  V = W_{V} \cdot \tau(c),
\end{align}
where $\varphi$ denote the flattening operation and $\tau$ is the condition embedding encoder and $W_Q, W_K$ and $ W_V$ is the learnable projection matrices. In this study, we adopt a cross-attention mechanism to integrate textual and visual cues, aligning with recent state-of-the-art ALDMs \cite{liu2023audioldm2,xue2024auffusion}.


\figframework

\subsection{\modelname}
\label{subsec:model}
\vspace{-2mm}
\modelname\ is a modular framework that can be integrated with a pre-trained text-to-audio (T2A) model, typically trained on large-scale high-quality audio datasets such as FreeSounds \cite{freesound_website,xue2024auffusion}. This design ensures both high-fidelity and diverse audio synthesis while allowing direct audio supervision for optimizing \modelname. As shown in \cref{fig:framework}, the framework consists of two main components: the \semanticmodule\ and the \temporalmodule, specifically designed for semantic alignment and temporal alignment, respectively. During training, only the \modelname\ components are trainable, optimizing with the supervision of ground truth audio, while the weights of the T2A model remain fixed. In the following sections, we provide more detailed explanations of each component.

\label{subsec:fol}


\figsemantic
\subsubsection{Semantic Adapter}
\label{subsubsection:sa}
\vspace{-2mm}
To efficiently extract semantic features from the input video and incorporate them into the pre-trained text-to-audio generator, we employ a visual encoder along with decoupled parallel cross-attention layers. 
We demonstrate the overview of the \semanticmodule\ in \cref{fig:semantic}.

\vspace{-2mm}
\paragraph{Visual Encoder.}
The CLIP encoder has demonstrated its effectiveness as a semantic extractor for visual information \cite{radford2021learning}. In our approach, we follow previous works \cite{rombach2022high,ye2023ip} and extract visual embeddings from each frame of the input video using the CLIP image encoder. To align these embeddings with the text-to-audio generator, we employ several learnable projection layers. This process can be expressed as:
\begin{equation}
V_{emb} = MLP(AvgPooling(\tau_{vis}(v))).
\end{equation}
Here, $v$ represents the input video, $\tau_{vis}$ denotes the CLIP image encoder, and $AvgPooling$ refers to the average pooling of the extracted CLIP features across frames.

\paragraph{Semantic Adapter.}
To incorporate the extracted video embedding into the pre-trained text-to-audio generator without compromising its original functionality, we introduce visual-conditioned cross-attention layers alongside the existing text-conditioned cross-attention layers. In this approach, visual and text embeddings are separately fed into their corresponding cross-attention layers. The outputs of the new and original cross-attention layers are then combined using a weight parameter, $\lambda$.
The parallel cross-attention can be denoted as:
\begin{align}
    Attention(Q, K, V) &= softmax(\frac{QK_{txt}^T}{\sqrt{d}}) \cdot V_{txt} + \lambda 
  \cdot softmax(\frac{QK^T_{vis}}{\sqrt{d}}) \cdot V_{vis}, \\
  \text{where } K_{txt} &= W_{K}^{txt} \cdot T_{emb}, V_{txt} = W_{V}^{txt} \cdot T_{emb},\\
   K_{vis} &= W_{K}^{vis} \cdot V_{emb}, V_{vis} = W_{V}^{vis} \cdot V_{emb},
\end{align}
where $T_{emb}$ and $V_{emb}$ represent the extracted text embeddings and video embeddings, respectively. $W_{K}^{txt}$ and $W_{V}^{txt}$ correspond to the pre-trained projection layers in the text-conditioned cross-attention layers, which remain fixed during training. On the other hand, $W_{K}^{vis}$ and $W_{V}^{vis}$ are newly introduced learnable projection layers used to map the visual embedding to a space that aligns better with the condition space of the pre-trained text-to-audio generator.

During the training of the \semanticmodule, we initialize the vision-conditioned cross-attention layers from the text-conditioned ones. As shown in \cref{fig:semantic}, we train the newly added projection layers after the visual encoder and the vision-conditioned cross-attention layers using ground truth audio as supervision. Meanwhile, we keep the text encoder and the text-to-audio generator fixed. The optimization objective can be expressed as:
\begin{equation}
    \mathcal{L} = \mathbb{E}_{x,\epsilon\sim N(0, 1), t, c} \left[\Vert \epsilon - \epsilon_{\theta}(z_t, t, T_{emb}, V_{emb}) \Vert \right]
    \label{equ:loss_sd1}.
\end{equation}
We observed a related work, IP-Adapter \cite{ye2023ip}, which also incorporates parallel cross-attention layers to inject image information into a pre-trained text-to-image generator. However, our \semanticmodule\ differs in that it introduces a third modality, namely video, into a text-to-audio generator. To effectively capture the visual cues and align them with the condition space of the text-to-audio generator, we randomly drop the text condition during training in the majority of cases (approximately 90\%). This design enables the \semanticmodule\ to effectively capture visual cues for audio generation while retaining the capability of combining text prompts for more controllable video-to-audio generation.



\vspace{-2mm}
\subsubsection{Temporal Controller}
\label{subsubsec:ta}
\vspace{-2mm}
We have observed that the \semanticmodule\ captures video-level alignment without precise temporal synchronization for each frame. To address this limitation, we have developed the \temporalmodule\ to enhance the temporal synchronization. 
As shown in \cref{fig:temporal}, the \temporalmodule\ utilizes a timestamp detector to predict sound and silence intervals within a given video. These predicted timestamp conditions are then fed into a temporal adapter, which controls the generation of audios.

\vspace{-2mm}
\paragraph{Timestamp Detector.}
The timestamp detector takes video frames as input and predicts a binary time condition mask, which indicates the presence or absence of sound effects in the target audio. We train the timestamp detector using binary cross-entropy loss, leveraging sound event labels. The training process can be expressed as:
\begin{equation}
\mathcal{L}_{BCE}(y, \hat{y}) = -\frac{1}{N} \sum_{i=1}^{N} \left[y_i \log(\hat{y}_i) + (1 - y_i) \log(1 - \hat{y}_i)\right]
\end{equation}
where $N$ represents the number of samples, $y$ denotes the ground truth timestamp label, and $\hat{y}$ represents the prediction. After training, the timestamp detector is capable of predicting the sound and silent timestamp mask, which is then passed to the temporal adapter for further processing.


\vspace{-2mm}
\paragraph{Timestamp-based Adapter.}
The timestamp-based temporal adapter shares the same architecture as the UNet encoder of the text-to-audio generator, following the design of ControlNet \cite{zhang2023adding}. Specifically, the temporal adapter utilizes the predicted timestamp condition mask to guide sound generation at the target timestamp. It takes the timestamp mask and the same latent input as the original UNet, and the output is then added as a residual to the output of the original UNet.
During training, we only train the replicated UNet blocks using the same optimization objective as the diffusion model. We train the temporal adapter by providing the timestamp labels of sound events from Audioset Strong \cite{hershey2021benefit} as input, with the corresponding audio as the target.

\vspace{-2mm}
\subsection{Implementation details}
\label{subsec:train}
\vspace{-2mm}
For the \semanticmodule, we follow \cite{ye2023ip} to use a linear projection for clip visual embedding to better align with text representation and expand the embedding length to four. Then we modify all the cross-attention to parallel cross-attention for visual conditions.
We train \semanticmodule\ on the VGGSound for 164 epochs with a batch size of 128 on 8 NVIDIA A100 GPUs. 
For the \temporalmodule, we train the timestamp detector and temporal adapter separately.
The timestamp detector is trained on the dataset with higher relevance audio-visual i.e. AVSync15 \cite{zhang2024audio} and Countix-AV \cite{zhang2021repetitive} for 23 epochs. The temporal adapter is trained on the AudioSet Strong \cite{hershey2021benefit} for 80 epochs.
Following, the two components in \temporalmodule can be combined together for inference and evaluation.



\figtemporal

%% file: sections/04_experiment.tex
\newcommand{\tabsemantic}{
\begin{table}
\caption{Quantitative evaluation in terms of semantic alignment and audio quality. Specifically, \modelname\ achieves state-of-the-art performance in terms of Mean KL Divergence (MKL) \cite{iashin2021taming}, CLIP \cite{wu2022wav2clip} and FID \cite{heusel2017gans} on standard benchmarks, \ie\ VGGSound \cite{chen2020vggsound} and AVSync15 \cite{zhang2024audio}. We report the results with error bars calculated from ten times of evaluation with random seeds.} 
\label{tab:sa}
\centering
\begin{tabular}{lccc}
\toprule[1.5pt]
\multicolumn{1}{c}{VGGSound \cite{chen2020vggsound}}  & MKL$\downarrow$     & CLIP$\uparrow$     & FID$\downarrow$          \\ 
\midrule[1.5pt]
SpecVQGAN(BN Inception)  \cite{iashin2021taming}   &       4.337$\pm 0.001$  & 5.079$\pm0.023$         &    65.37$\pm 0.01$      \\
SpecVQGAN(ResNet)    \cite{iashin2021taming}       &        3.400$\pm 0.001$     & 5.876$\pm0.016$          &      32.01$\pm 0.005$            \\
Diff-Foley  \cite{luo2024diff}                &     3.318$\pm 0.011$   & 9.172$\pm0.110$      &    29.03$\pm 0.61$       \\
V2A-Mapper \cite{wang2024v2a} &        2.654              & 9.720       &  24.16          \\
\modelname\ (ours)                        &         \textbf{2.561$\pm 0.011$}        & \textbf{10.70$\pm 0.121$}   & \textbf{19.67$\pm 0.05$}     \\ 

\midrule[1.5pt]
\multicolumn{1}{c}{AVSync15 \cite{zhang2024audio}} & \multicolumn{1}{c}{MKL  $\downarrow$} & \multicolumn{1}{c}{CLIP  $\uparrow$} & \multicolumn{1}{c}{FID  $\downarrow$}   \\ 
\midrule[1.5pt]
SpecVQGAN(Inception) \cite{iashin2021taming}    &      5.339$\pm0.077$         &    6.610$\pm0.014$      &  114.44$\pm1.31$       \\
SpecVQGAN(ResNet)   \cite{iashin2021taming}        &3.603$\pm0.006$                  & 6.474$\pm0.077$    & 75.56$\pm1.43$\\
Diff-Foley     \cite{luo2024diff}             &    1.963$\pm0.007$                    & 10.38$\pm0.008$  &     65.77$\pm0.01$          \\
Seeing and Hearing     \cite{xing2024seeing}             &    2.547                      & 2.033  &      65.82                                          \\
\modelname\ (ours)     & \textbf{1.497$\pm0.006$}         & \textbf{11.94$\pm0.217$}  &   \textbf{36.80$\pm2.67$}
\\\bottomrule[1.5pt]
\end{tabular}
\end{table}
}

\newcommand{\tabsync}{
\begin{table}
\caption{Quantitative evaluation in terms of temporal synchronization. We report onset detection accuracy (Onset ACC) and average precision (Onset AP) for the generated audios on AVSync \cite{zhang2024audio}, which provides onset timestamp labels for assessment, following previous studies \cite{luo2024diff,xie2024sonicvisionlm}. We report the results with error bars calculated from ten times of evaluation with random seeds.}
\label{tab:temp}
\centering
\begin{tabular}{lcc}
\toprule[1.5pt]
\multicolumn{1}{c}{Method} & \multicolumn{1}{c}{Onset ACC $\uparrow$} & \multicolumn{1}{c}{Onset AP  $\uparrow$}  \\ 
\midrule[1.5pt]
SpecVQGAN(Inception) \cite{iashin2021taming}    & 16.81$\pm1.14$                              & 64.64$\pm4.28$                                       \\
SpecVQGAN(ResNet)   \cite{iashin2021taming}        & 26.74$\pm2.35$                  & 63.18$\pm0.72$                                      \\
Diff-Foley     \cite{luo2024diff}             & 21.18$\pm0.08$                       & 66.55$\pm0.10$                       \\
Seeing and Hearing     \cite{xing2024seeing}             &    20.95                      & 60.33                         \\
\modelname\ (ours)        & \textbf{28.48$\pm1.84$}                           & \textbf{68.14$\pm1.03$}                                   \\
\bottomrule[1.5pt]
\end{tabular}
\end{table}
}


\newcommand{\tababa}{
\begin{wraptable}{h}{0.50\linewidth}
\vspace{-4mm}
\caption{Ablation on \semanticmodule.}
\label{table:aba}
\centering
\begin{tabular}{ccccc}
\toprule[1.5pt]
 \multicolumn{1}{c}{Method} & \multicolumn{1}{c}{MKL$\downarrow$} & \multicolumn{1}{c}{CLIP Score$\uparrow$ } & \multicolumn{1}{c}{FID$\downarrow$}      \\ \midrule[1.5pt] 
Captioner   &      2.331                   &9.177       &               67.40               \\
Visual      & 5.383 &2.133            &         99.77                        \\
Visual*    & 5.821 &2.778          &     95.78                           \\
Ours  & \textbf{1.497}                         & \textbf{11.94}                             & \textbf{36.80}                   \\
\bottomrule[1.5pt]
\end{tabular}
\vspace{-2mm}
\end{wraptable}
}

\newcommand{\figqual}{
\begin{figure}[t]
    \centering
    \includegraphics[width=\textwidth]{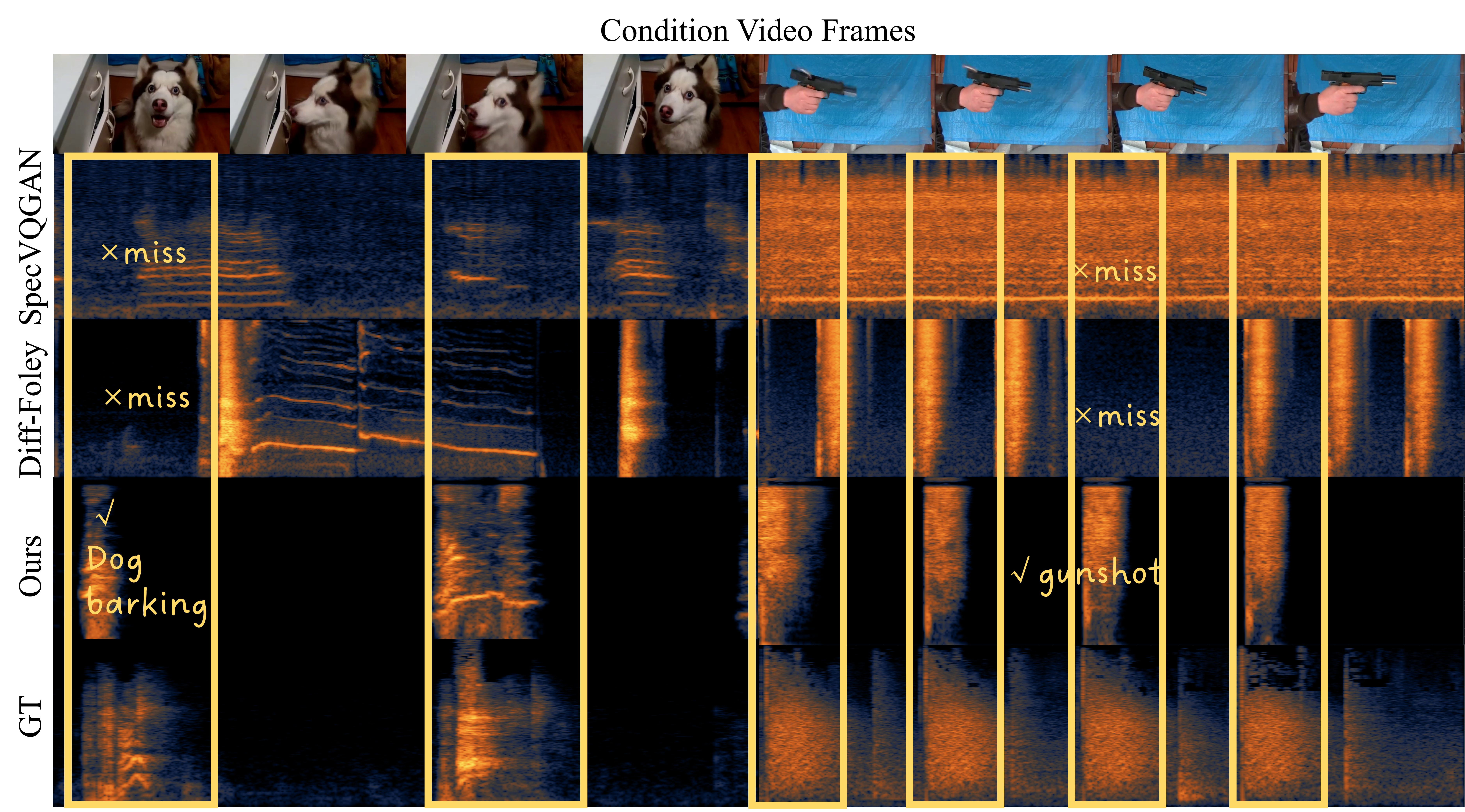}
    \caption{\textbf{Qualitative comparison.} 
     As shown in the first case, both SpecVQGAN and Diff-Foley fail to capture the onset of the gunshot sound. In contrast, \modelname\ generates the gunshot sound synchronized with the video, showcasing its superior temporal alignment capability.
    }
    \label{fig:temporal_comparison}
\vspace{-3mm}
\end{figure}
}

\newcommand{\figprompt}{
\begin{figure}[t]
    \centering
    \includegraphics[width=\textwidth]{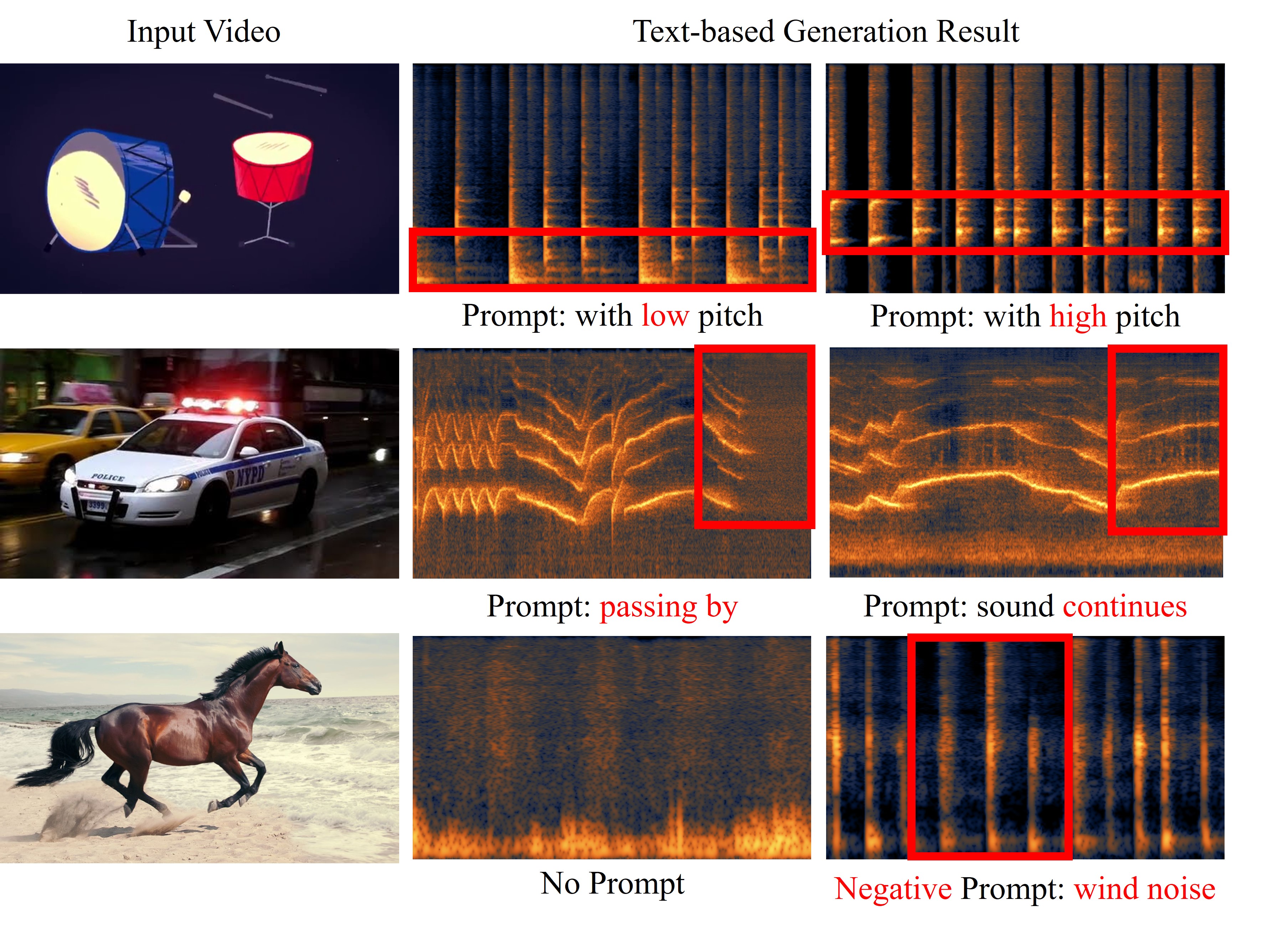}
    \vspace{-6mm}
    \caption{
    \textbf{Video-to-audio generation with text prompts.} 
   \modelname\ enhances controllability in video-to-audio generation through text prompts. In the first case, providing a prompt for "high pitch" increases the corresponding value for the drum video. In the third case, a negative prompt like "wind noise" can be used during inference to prevent the generation of wind noise for the horse video.
   }
    \label{fig:prompt_control}
    \vspace{-3mm}
\end{figure}
}


\section{Experiments}
\label{sec:exp}
\subsection{Experimental settings}
\paragraph{Baselines.}
We conducted comprehensive evaluations of \modelname\ by comparing it with state-of-the-art approaches, namely SpecVQGAN \cite{iashin2021taming}, Diff-Foley \cite{luo2024diff}, V2A-Mapper \cite{wang2024v2a}, and Seeing-and-hearing \cite{xing2024seeing}. Both quantitative and qualitative comparisons were employed. SpecVQGAN generates audio tokens autoregressively based on extracted video tokens. Diff-Foley utilizes contrastive learning for synchronized video-to-audio synthesis with its CAVP audio and visual encoder. V2A-Mapper translates visual CLIP embeddings to CLAP space, enabling video-aligned audio generation using a pre-trained text-to-audio generator. Seeing-and-hearing \cite{xing2024seeing} propose using ImageBind \cite{girdhar2023imagebind} as a bridge between visual and audio, leveraging off-the-shelf audio and video generators for multimodal generation. Please note that the results of V2A-Mapper and Seeing-and-hearing were obtained directly from the authors, as their source codes and models are not publicly available at this time.

\paragraph{Evaluation Metrics.}
We employed several evaluation metrics to assess semantic alignment and audio quality, namely Mean KL Divergence (MKL) \cite{iashin2021taming}, CLIP similarity, and Frechet Distance (FID) \cite{heusel2017gans}, following the methodology of previous studies \cite{luo2024diff,wang2024v2a,xing2024seeing}. MKL measures paired sample-level similarity by calculating the mean KL-divergence across all classes in the test set. CLIP Score compares the similarity between the input video and the generated audio embeddings in the same representation space. For this, we employed Wav2CLIP \cite{wu2022wav2clip} as the audio encoder and CLIP \cite{radford2021learning} as the video encoder, as done in previous works \cite{wang2024v2a, sheffer2023hear}. FID assesses the distribution similarity to evaluate the fidelity of the generated audio.
Additionally, we follow Du et al. \cite{du2023conditional, xie2024sonicvisionlm} and adopt onset detection accuracy (Onset Acc) and onset detection average precision (Onset AP) to evaluate the generated audios, using the onset ground truth from the datasets.


\tabsemantic
\tabsync
\vspace{-2mm}
\subsection{Comparison with State-of-the-art}
\vspace{-2mm}
\paragraph{Quantitative Comparison.}
We present a quantitative comparison of semantic alignment and audio quality on both the VGGSound \cite{chen2020vggsound} and AVSync15 \cite{zhang2024audio} datasets, as shown in \cref{tab:sa}. The VGGSound dataset consists of 15,446 videos sourced from YouTube, encompassing a wide range of genres. The results indicate that \modelname\ achieves superior semantic alignment with visual conditions and provides higher audio fidelity.
Furthermore, we report the results for temporal synchronization on the AVSync15 dataset \cite{zhang2024audio}, as displayed in \cref{tab:temp}. The AVSync15 dataset is a carefully curated collection of video-audio pairs with strong video-audio alignment and onset detection labels. This characteristic makes it a reliable benchmark for evaluating synchronization. The results in \cref{tab:temp} demonstrate that when combined with the \temporalmodule, \modelname\ achieves state-of-the-art performance in terms of temporal synchronization.


\figqual
\vspace{-2mm}
\paragraph{Qualitative Comparison.}
We provide the visualization of generated audio for qualitative comparison on the AVSync15 \cite{zhang2024audio} in \cref{fig:temporal_comparison}. It can be observed that \modelname\ generates sound at the most accurate time aligned with visual cues, closely resembling the pattern of the ground truth audio. 
However, SpecVQGAN tends to introduce more noise, while Diff-Foley often generates more or fewer sound events compared to the ground truth. We provide more results in the Appendix.

One notable advantage of \modelname\ is its compatibility with text prompts, allowing for more controllable Foley. We present visualization results of audio generation conditioned on both a video and a text prompt in \cref{fig:prompt_control}.
For instance, when the text prompt describes "high pitch," the corresponding value for high-frequency increases compared to when the prompt describes "low pitch." 
Moreover, \modelname\ can also be utilized with negative prompts to prevent the generation of unwanted sounds. In the third case shown in \cref{fig:prompt_control}, the visual cues depict a horse running on the beach. By setting the negative prompt as "wind and noise" during inference, the generated audio successfully removes the sound of wind and other environmental noise, resulting in a clear sound of hooves. 
We provide more results in the Appendix.


\vspace{-2mm}
\subsection{Ablation Study}
\label{subsec:abl}
\vspace{-2mm}
\tababa
We conduct ablation studies to validate the effectiveness of \semanticmodule\ and \temporalmodule.
For \semanticmodule, we compare the audio-visual relevance of generated samples using different methods of video information injection, as shown in \cref{table:aba}. We consider several baselines for comparison. First, we use a captioner model that utilizes a video-text captioning model \cite{achiam2023gpt} to generate text descriptions as inputs to the text-to-audio generator. Second, we directly feed the visual embedding into cross-attention as the text prompt embedding, without any training. Third, we fine-tune the cross-attention module to adapt it to the visual embedding (referred to as `visual*' in \cref{table:aba}).
We observed that the caption-based method struggles to capture all the details in the video, resulting in sub-optimal generation results with visual captioning. Using the visual embedding with or without fine-tuning UNet both fail to generate audio that is relevant to the input video. We attribute this to the significant distortion of the original text-to-audio generation framework when incorporating visual information.

For \temporalmodule, we compare the temporal synchronization performance of \modelname\ with and without the module. The results in \cref{table:abb} demonstrate that the absence of the \temporalmodule\ leads to a noticeable decline in onset precision.
This decline can be attributed to the fact that the \semanticmodule\ is only capable of capturing video-level semantic information without accurate synchronization features. As a result, it tends to synthesize relevant sounds but with random onset timestamps, leading to a lack of precise temporal alignment.

\figprompt

%% file: sections/05_conclusion.tex
\newcommand{\tababb}{
\begin{wraptable}{r}{0.50\linewidth}
\vspace{-5mm}
\caption{Ablation on \temporalmodule.}
\label{table:abb}
\centering
\begin{tabular}{ccc}
\toprule[1.5pt]
 \multicolumn{1}{c}{Method} & \multicolumn{1}{c}{Onset Acc$\uparrow$ } & \multicolumn{1}{c}{Onset AP$\uparrow$ } \\ \midrule[1.5pt] 
\semanticmodule\          & 26.65 & 63.20 \\
\modelname  & \textbf{28.48}  & \textbf{68.14} \\
\bottomrule[1.5pt]
\end{tabular}
\vspace{-3mm}
\end{wraptable}
}

\vspace{-3mm}
\section{Limitations and Broader Impatct}
\vspace{-3mm}
\tababb
\paragraph{Limitations.} 
Firstly, although the inclusion of the \temporalmodule\ enhances the synchronization between the generated audio and the input video, its performance can be ultimately limited by the capabilities of the temporal detector. 
Second, the effectiveness of the temporal detector is contingent upon the availability of strong and relevant training data. 
When dealing with more complex visual scenes, predicting the onset event for accurate synchronization becomes challenging due to the scarcity of training data in those specific contexts.

\vspace{-3mm}
\paragraph{Broader Impact.}
\modelname\ facilitates text-based video-to-audio generation, enabling the generation of sound effects for silent videos and providing control through user prompts. However, it is crucial to acknowledge the potential misuse of such technology for generating fake content on video platforms or social platforms. Users and researchers are strongly advised to exercise caution and carefully screen the use of such technologies to ensure responsible and ethical application.


\vspace{-3mm}
\section{Conclusion}
\vspace{-3mm}
In this paper, we introduce \modelname\ for adding sound effects to silent videos. Unlike existing methods that either train a video-to-audio generator from scratch or use video-to-text translation followed by text-to-audio generation, \modelname\ is a pluggable module seamlessly integrated into a text-to-audio generator. This integration ensures high-quality audio generation while synchronizing with the video content. \modelname\ leverages two key components, namely \semanticmodule\ for semantic alignment and \temporalmodule\ for temporal synchronization. Extensive experiments on standard benchmarks demonstrate the effectiveness of \modelname.

\section{Acknowledgement}
This project is supported by the National Key R\&D Program of China (No. 2022ZD0161600) and partially supported by NSFC (project 62376237) and Shenzhen Science and Technology Program ZDSYS20230626091302006.



%% file: sections/06_appendix.tex
\newcommand{\tabuser}{
\begin{table}
\caption{\textbf{User study.} We evaluated the performance of three metrics of different models i.e. semantic and temporal alignment and generation quality.}
\centering
\begin{tabular}{lccc}
\toprule[1.5pt]
\multicolumn{1}{c}{Method} & \multicolumn{1}{c}{Semantic} & \multicolumn{1}{c}{Temporal} & \multicolumn{1}{c}{Quality} \\ 
\midrule[1.5pt]
SpecVQGAN                  &20.29                             &21.74                              &20.29                             \\
Diff-Foley                 &20.59                              &29.41                              &27.94                             \\
V2A-Mapper                 &44.00                              &44.00                              &42.67                             \\
FoleyCrafter (ours)                 &\textbf{71.23}                              &\textbf{67.92}                              &\textbf{69.34}
\\
\bottomrule[1.5pt]
\label{table:ablation}
\end{tabular}
\end{table}
}

\newcommand{\figuser}{
\begin{figure}[t]
    \centering
    \includegraphics[width=\textwidth]{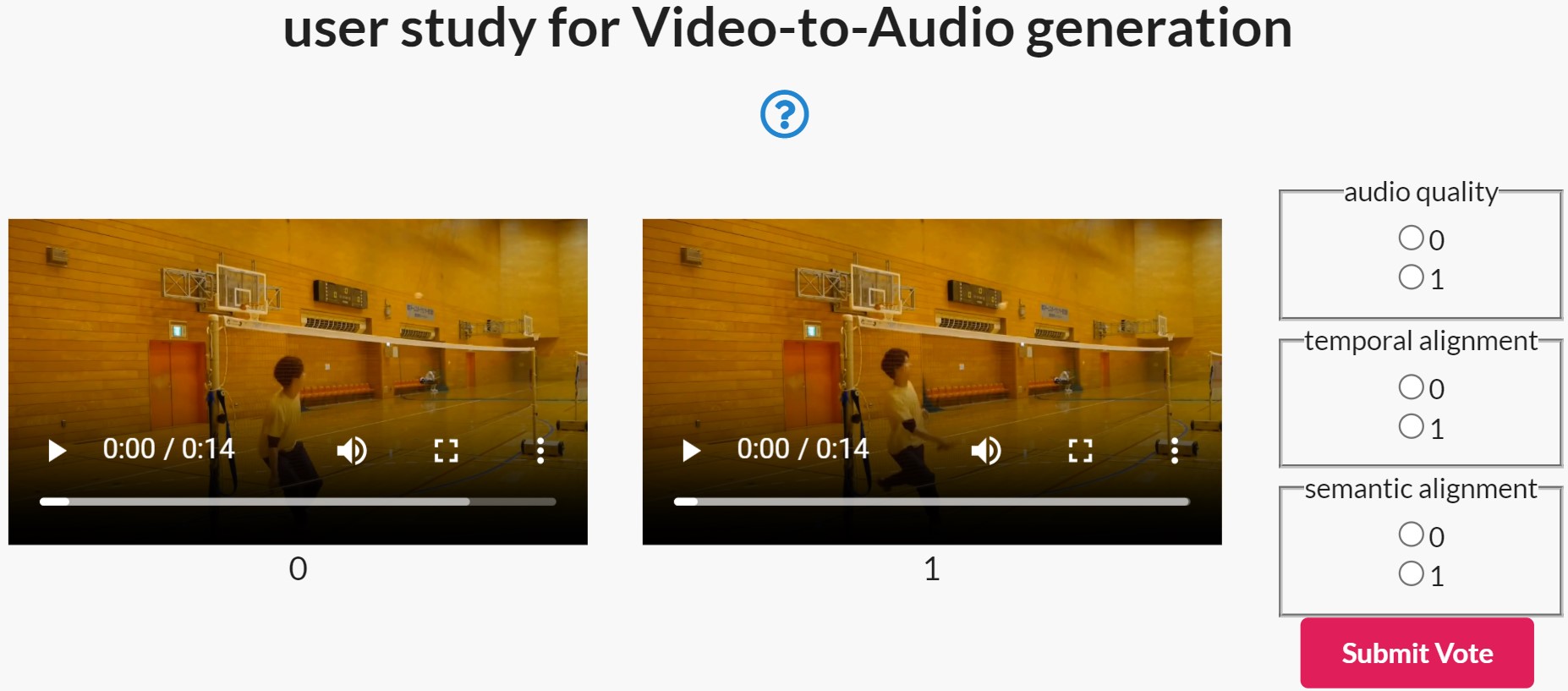}
    \caption{\textbf{Screenshot of User Study.}}
    \label{fig:user}
\end{figure}
}

\newcommand{\figuserextra}{
\begin{figure}[ht]
    \centering
    \includegraphics[width=0.85\textwidth]{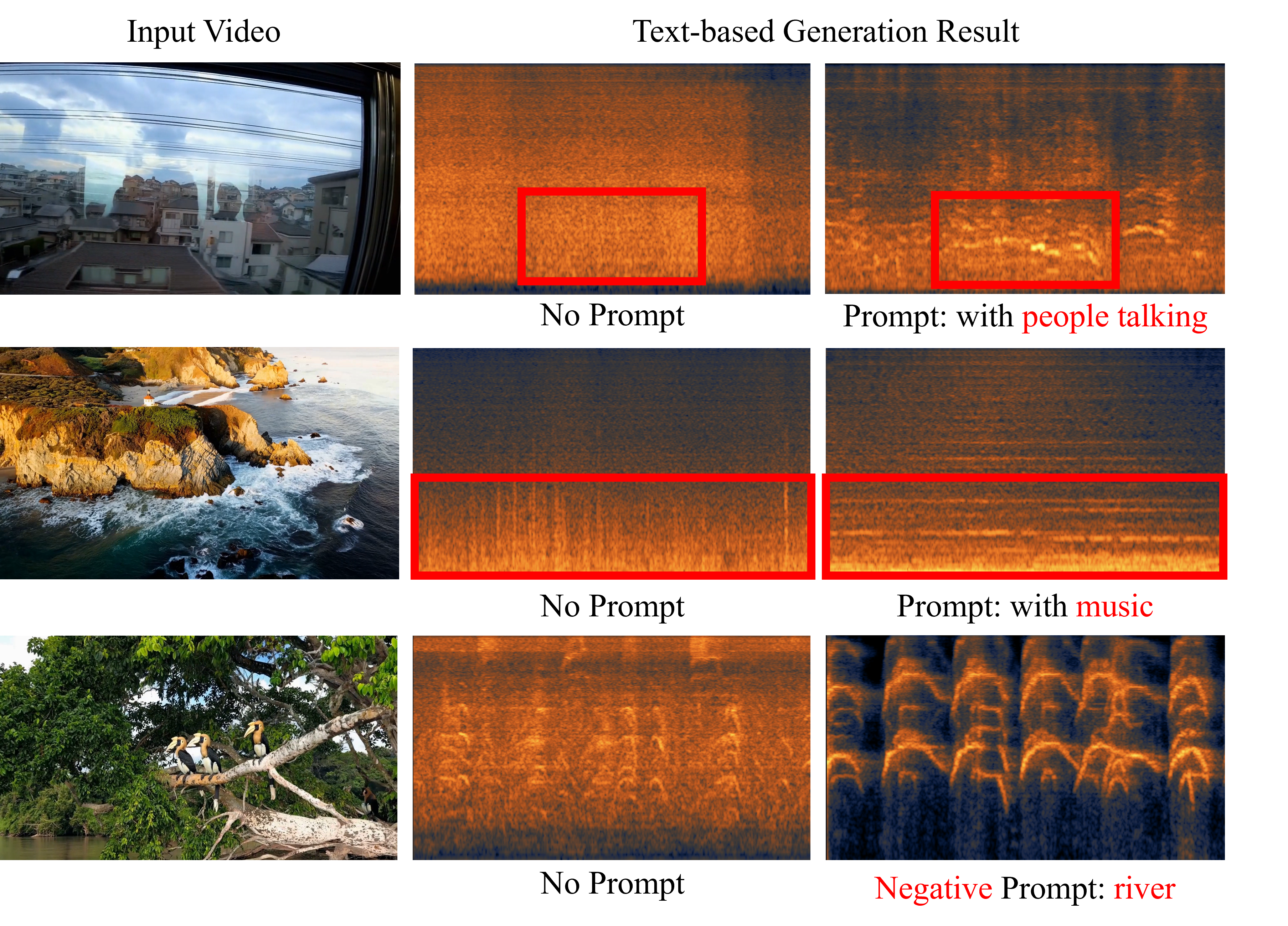}
    \vspace{-5mm}
    \caption{\textbf{Extra examples on text-based video to audio generation.}}
    \label{fig:edit}
    \centering
    \hspace{-0.875cm}\includegraphics[width=0.85\textwidth]{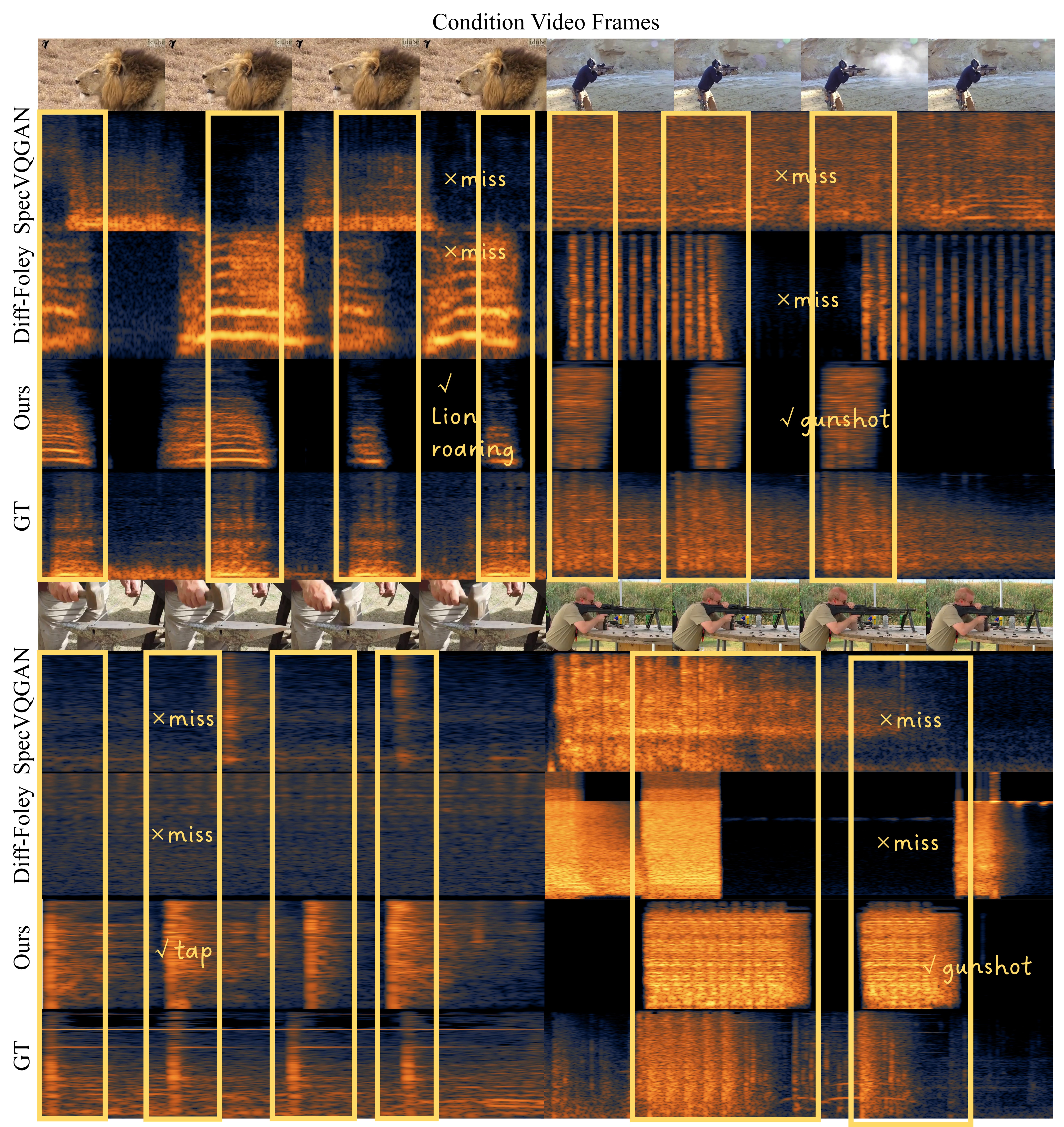}
    \vspace{-3mm}
    \caption{\textbf{Extra examples on temporal alignment comparison.}}
    \label{fig:comparison}
\end{figure}
}

    


\section{Appendix}

\noindent
\textbf{Overview.} The appendix includes the following sections:
\begin{itemize}
    \item \textbf{\cref{sec:supp-training-datasets}.} Details of training datasets.
    \item \textbf{\cref{sec:user-study}.} Details of the user study.
    \item \textbf{\cref{sec:supp-qualitative-results}.} More qualitative results.
\end{itemize}

\noindent
\textbf{Video Result.} We also present video results in a separate supplementary file sourced from Sora.

\subsection{Details of Training Dataset}
\label{sec:supp-training-datasets}
\modelname\ consist of two key components: \semanticmodule\ and \temporalmodule\ which are trained separately. For the training of \semanticmodule\, we use VGGSound \cite{chen2020vggsound} as the training set. VGGSound is an audio-visual dataset containing approximately 199,176 videos sourced from YouTube with annotated label classes indicating the video content. We add the prefix 'The sound of' to the label to form the prompt for generation. 
We train the timestamp detector on the AVSync15 \cite{zhang2024audio}. AVSync15 is a carefully curated dataset from the VGGSound Sync \cite{chen2021audio} dataset, which contains 1500 strongly correlated audio-visual pairs, making it a high-quality dataset for timestamp detection.
We use the AudioSet Strong \cite{hershey2021benefit} to train the temporal adapter, which contains 103,463 videos with the sound event and the corresponding timestamp labels.



\subsection{Details of user study}
\label{sec:user-study}

To further obtain subjective evaluation results, we conduct a user study. We randomly selected the VGGSound test results generated by different methods for the questionnaire. A total of 20 participants answered our questions.  
As shown in \cref{fig:user}, each question contains audios generated by two methods, one is our method and the other is the baseline e.g. SpecVQGAN \cite{iashin2021taming} Diff-Foley \cite{luo2024diff} and V2A-Mapper \cite{wang2024v2a}.
We ask participants to select the one that has better semantic alignment, temporal alignment, and generation quality.
Then the preference score can be calculated as
\begin{equation}
    Score = \frac{S}{A}
\end{equation}
where $S$ is the number of times the method has been selected and $A$ is the appearance times of that method. 
A higher score means the greater performance of \modelname.
Results can be found at \cref{table:ablation}. \modelname\ is preferred by users in all three metrics. 

\figuser
\tabuser

\subsection{More qualitative results}
\label{sec:supp-qualitative-results}

\paragraph{Foley Generation for Generated Videos.}
\modelname\ is an effective Foley generation tool which can also be used for movie and generated video. Herein, we take the Sora video as example and provide the audio results generated by \modelname. In the foley process, \semanticmodule\ can directly utilize the rich visual information, which helps \modelname\ generate appropriate sound effects for the visual subjects and environment shown in the generated videos.  

\figuserextra

\paragraph{Text-based video to audio generation.}
\modelname\ achieve text-based video-to-audio generation through parallel cross-attention in \semanticmodule. Benefiting from this module, \modelname\ can utilize both visual information and text prompts to generate audio. Extra text-based video-to-audio generation results are illustrated in \cref{fig:edit} and attached in a separate supplementary file.

\paragraph{Temporal Synchronization Comparison.}
The temporal controller enhances the temporal alignment in generated audios with visual cues. To show the synchronization ability of \modelname, we show more intuitive comparison results between \modelname\ and other methods as shown in \cref{fig:comparison}. Video results are also provided in a separate supplementary file.

\paragraph{Video to Audio generation on various genres.}
\modelname\ can generate audio for a wide variety of videos. In the supplementary file, we provide generated audio-visual pairs from the VGGSound test cases. The type of video contains realistic video, games, and animation. The main visual objects in the video are people, animals, musical instruments, etc. It fully demonstrates the excellent video-to-audio generation capabilities of \modelname.

\clearpage